\PassOptionsToPackage{hyphens}{url}
\documentclass[journal]{IEEEtran}

\usepackage{url}
\usepackage{graphicx}   
\usepackage{subcaption} 
\usepackage{tabularx}   
\usepackage{float}      
\usepackage{pgfplots}   
\usepackage{tikz}       
\pgfplotsset{compat=1.18}
\usepackage{inconsolata}
\usepackage{hyperref}
\usepackage{amsmath}
\usepackage{amsfonts}

\begin{document}

\title{REMIND: RE-Identification with Memory for INDoor Navigation}

\author{Pablo~Diaz-Pereda,
        Alejandro~Rodriguez-Ramos,
        David~Perez-Saura,
        and~Pascual~Campoy
\thanks{All authors are with the Computer Vision and Aerial Robotics Group at Centre for Automation and Robotics C.A.R. (UPM-CSIC), Universidad Politécnica de Madrid (CVAR-UPM), Calle Jose Gutierrez Abascal 2, 28006 Madrid, Spain.}}

\maketitle

\begin{abstract}
Mobile robots operating indoors must re-identify previously observed objects after long temporal gaps, significant viewpoint changes, and severe illumination variations. This remains a challenging problem: multi-object tracking methods are optimized for short-term association of pedestrians and vehicles at video rates, person and vehicle re-identification approaches lack persistent memory mechanisms, and state-of-the-art video object segmentation techniques rely on reactive distractor filtering rather than enforcing global identity consistency. 

To address these limitations, we present REMIND, an online tracker designed for long-term multi-object re-identification of generic indoor objects from monocular RGB imagery, requiring neither camera pose nor depth. Motivated by evidence from visual cognition that humans rely on accumulated appearance familiarity and spatial context rather than explicit self-localization, REMIND combines frozen DINOv3 features with a dual-bank multi-prototype appearance memory, part- and background-level descriptors, a neighbour-context reasoning module exploiting spatial co-occurrence, and joint Hungarian assignment with ambiguity-aware safeguards. On a purpose-built indoor dataset featuring controlled revisits and dense same-class clutter, REMIND reaches 90.35\% IDF1, nearly 20 points above a state-of-the-art video object segmentation baseline and more than 36 above a strong tracking-by-detection baseline. On ScanNet++, it attains the highest IDF1 in every setting but one, end-to-end detection over all scenes, where the tracking-by-detection baseline is marginally ahead while REMIND still associates and recovers identities more accurately; it also completes every scene, whereas the video object segmentation baseline exhausts GPU memory on 66.9\% under YOLO detections. The complete system, evaluation framework, and dataset are \href{https://cvar-vision-dl.github.io/remind-reid-tracker/}{publicly released}.

\end{abstract}


\IEEEpeerreviewmaketitle

\section{Introduction}

Humans recognise previously seen objects with remarkable ease, even after long
absences and under substantially different viewing conditions. Returning to a
room after days, a person re-identifies familiar items despite changes in
viewpoint, illumination, and partial occlusion, without any conscious
estimation of their own position in space. Research in visual cognition has
shown that this ability relies heavily on accumulated appearance familiarity and
on the spatial relationships between co-occurring
objects~\cite{oliva2007context}, rather than on explicit geometric
reconstruction of the scene. The robustness of this process, and its
independence from self-localisation, points to a computational principle
directly relevant to autonomous systems that must maintain object identity over
time.

\begin{figure}[!t]
  \centering
  \includegraphics[width=\columnwidth]{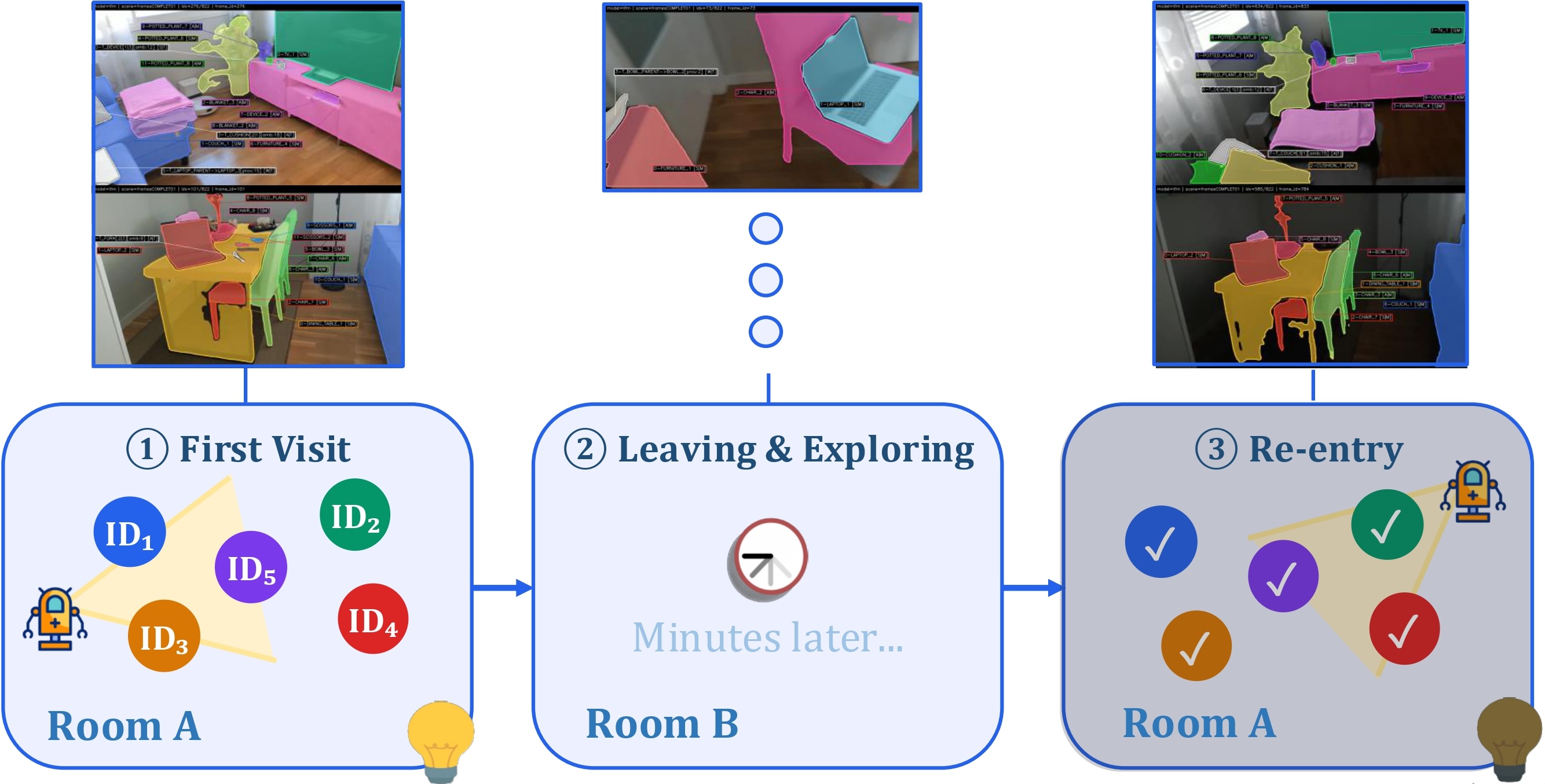}
  \caption{Long-term object re-identification scenario for indoor robot
  navigation. \textbf{(1)~First visit:} the robot enters Room~A, perceives
  the scene, and assigns persistent identities
  $\mathrm{ID}_1,\ldots,\mathrm{ID}_5$ to the visible objects.
  \textbf{(2)~Leaving and exploring:} the robot traverses other rooms for
  several minutes, during which the previously observed objects are out of
  view. \textbf{(3)~Re-entry:} upon returning to Room~A, every previously
  catalogued object must be re-identified despite substantial changes in
  viewpoint, illumination, and partial occlusion, and without relying on
  camera pose or scene geometry.}
  \label{fig:overview}
\end{figure}

Mobile robots navigating indoor environments face precisely this challenge. A
robot leaves a room, traverses a corridor, and returns minutes later; it must
re-identify the same objects under substantial viewpoint and illumination
changes (Fig.~\ref{fig:overview}). Unlike frame-to-frame tracking, where
objects move smoothly through consecutive video frames, this setting demands
\emph{long-term identity persistence}: the ability to bridge temporal gaps of
hundreds of frames and recover object identities after complete disappearance
and re-appearance. Moreover, the objects of interest are not restricted to
well-studied categories such as pedestrians or vehicles; they span the full
diversity of generic indoor items (chairs, backpacks, monitors, plants) for
which no category-specific model exists.
 
The aforementioned human's cognitive behavior carries a practical implication for tracking systems design:
accurate camera pose is not strictly necessary for object re-identification. Since
humans reliably recognise objects without estimating their own pose, an
appearance-and-context-based approach should be sufficient for the same task in
robotic settings. This motivates a system that operates \emph{entirely from
monocular RGB}, without camera pose, depth sensing, or 3D reconstruction,
relying on appearance memory and contextual reasoning as the primary
computational primitives.

Existing paradigms each address a subset of these requirements, but none was
designed for all of them simultaneously. Multi-object tracking
(MOT)~\cite{wojke2017simple, zhang2022bytetrack, du2023strongsort,
maggiolino2023deep, gao2023memotr} decomposes video into per-frame detections
followed by short-horizon data association, with re-identification modules
calibrated for inter-frame gaps at conventional video rates. Tracks are
terminated at scene exits, and both architectures and training data are 
specific to pedestrians and vehicles. Person and vehicle
re-identification~\cite{zheng2015scalable, wei2018person, liu2016deep,
he2021transreid, luo2019bag} provides powerful appearance models for
cross-camera matching, but these models are category-specific, operate in a
retrieval paradigm without temporal memory, and lack multi-object assignment.
Generic instance association methods such as MASA~\cite{li2024masa} achieve
class-agnostic matching across diverse categories, yet operate strictly
frame-to-frame without accumulating appearance evidence over time or
incorporating spatial context.
 
The closest functional comparison comes from the video object segmentation
(VOS) literature, particularly the SAM2 ecosystem~\cite{ravi2024sam2}.
SAM2Long~\cite{ding2025sam2longenhancingsam2} and DAM4SAM~\cite{videnovic2025CVPR} extend
SAM2's memory architecture to handle object disappearance and reappearance,
and DAM4SAM further introduces distractor-aware memory updates that suppress
frames contaminated by other tracked objects. This constitutes a limited form
of cross-object awareness, but it is reactive rather than proactive: the
mechanism filters out interference from nearby objects rather than jointly
reasoning about the identity assignment across all candidates. There is no
global assignment step that enforces identity consistency, making swaps between
visually similar objects an inherent failure mode. Furthermore, VOS memory
banks are designed for continuous video in which objects may briefly leave the
frame; their memory management strategies (frame selection, update frequency)
are not optimised for scenarios where an object disappears for hundreds of
frames and reappears under a substantially different viewpoint, as is common
when a robot re-enters a previously visited space.
 
In this work, we present REMIND (\textbf{RE}-identification with
\textbf{M}emory for \textbf{IND}oor Navigation), a system purpose-built for
long-term multi-object re-identification of generic indoor objects from
monocular video. Drawing on the principle that appearance memory and contextual
reasoning, rather than geometric reconstruction, are sufficient for robust
object recognition, REMIND combines foundation-model appearance features
(DINOv3~\cite{simeoni2025dinov3}) with an accumulated dual-bank object memory,
neighbour-context reasoning for disambiguation, and joint multi-object
assignment via the Hungarian algorithm, all operating on automatic per-frame
detections without manual prompts or camera pose information.
 
\medskip
\noindent\textit{Contributions:}
\begin{enumerate}
    \item Long-term multi-object re-identification of generic indoor
    objects from monocular video is identified as an underexplored
    problem setting at the intersection of MOT, Re-ID, generic
    instance association, and VOS, and a structured analysis is
    provided of why each paradigm's assumptions leave gaps for this
    setting.

    \item A re-identification tracker is introduced that combines
    accumulated appearance memory, neighbour-context reasoning, and
    joint multi-object assignment. On the custom indoor sequence with
    controlled re-entry scenarios, REMIND outperforms
    DAM4SAM~\cite{videnovic2025CVPR} by nearly 20 IDF1 points (90.35\%
    vs.\ 70.53\%) and MASA~\cite{li2024masa} by more than 36. On
    ScanNet++~\cite{yeshwanth2023scannetpp}, it attains the highest
    IDF1 in every regime except end-to-end detection over all scenes,
    where MASA is marginally ahead while REMIND retains higher
    association accuracy and recovery, and it completes 100\% of scenes
    against DAM4SAM's 66.9\% out-of-memory failure rate under YOLO
    detections.
    
    \item The full system, evaluation code, and custom dataset are
    released at
    \texttt{\url{https://cvar-vision-dl.github.io/remind-reid-tracker/}},
    providing the first dedicated baseline and evaluation protocol for
    this problem setting.
\end{enumerate}

\section{Related Work}

\subsection{Multi-Object Tracking}

Associating object identities across frames in video is the central problem of multi-object tracking (MOT).
Modern methods such as DeepSORT~\cite{wojke2017simple}, ByteTrack~\cite{zhang2022bytetrack}, StrongSORT~\cite{du2023strongsort}, and Deep OC-SORT~\cite{maggiolino2023deep} decompose the problem into per-frame detection followed by short-horizon data association, relying on Kalman filtering and appearance re-identification modules operating within windows of tens of frames.
More recent approaches such as MeMOTR~\cite{gao2023memotr} incorporate learned temporal memory to extend association horizons, but require end-to-end training on annotated tracking datasets and remain bound to the pedestrian and vehicle domains for which they were designed.
Critically, all of these methods assume temporally continuous object motion relative to a largely static camera, and their re-identification modules are calibrated for the inter-frame gaps characteristic of conventional video frame rates, not the minute-scale gaps that arise when a robot exits and re-enters a room.
In contrast, our approach is designed precisely for this setting: objects are static, the camera moves, gaps span hundreds of frames, and the tracked entities are generic indoor objects (requirements for which no existing MOT method was purpose-built).

\subsection{Person and Vehicle Re-Identification}

Person and vehicle re-identification (Re-ID) address the problem of matching identities across multiple cameras or large temporal gaps, making them a natural reference point for our setting.
Standard benchmarks such as Market-1501~\cite{zheng2015scalable}, MSMT17~\cite{wei2018person}, and VeRi-776~\cite{liu2016deep} have driven the development of powerful appearance models based on metric learning, attention mechanisms, and strong data augmentation strategies, achieving robust cross-view matching under viewpoint and illumination changes~\cite{he2021transreid, luo2019bag}.
However, these architectures and loss functions are fundamentally designed around the visual characteristics of human bodies and vehicles; they do not generalise to the diverse appearance space of generic indoor objects, such as chairs, backpacks, or monitors, without category-specific retraining.
Furthermore, Re-ID methods operate in a retrieval paradigm, querying a gallery for the closest match, rather than maintaining a temporal memory that accumulates evidence across observations or performing joint multi-object assignment across all active identities.
Our approach addresses both of these limitations by employing class-agnostic foundation-model features and an accumulated appearance memory that supports joint assignment over all tracked identities simultaneously.

\subsection{Generic Instance Association}

A key step toward class-agnostic tracking was taken by MASA~\cite{li2024masa}, which trains an instance-matching module on proposals generated by SAM~\cite{kirillov2023sam}, achieving zero-shot generalisation across over 800 object categories on the TAO benchmark~\cite{dave2020tao} without category-specific supervision.
This demonstrates that strong instance-level appearance representations can be learned without domain-specific annotations, establishing a compelling foundation for generic object tracking.
However, MASA operates strictly in a frame-to-frame association paradigm: it computes pairwise similarity between adjacent frames without accumulating appearance evidence over time or incorporating spatial context from co-visible neighboring objects.
As a result, performance degrades substantially under long temporal gaps and strong viewpoint changes, precisely the conditions that define our target setting.
Our approach builds upon the class-agnostic matching insight, but extends it with an accumulated object memory and neighbor-context reasoning that enable robust re-identification after gaps of hundreds of frames.

\subsection{Video Object Segmentation and the SAM2 Ecosystem}

The closest functional comparison to our techniques comes from the video object segmentation (VOS) literature, particularly methods built on SAM2~\cite{ravi2024sam2}.
SAM2Long~\cite{ding2025sam2longenhancingsam2} and DAM4SAM~\cite{videnovic2025CVPR} extend SAM2's frame-level memory architecture to handle object disappearance and reappearance in long sequences; DAM4SAM further introduces distractor-aware memory update rules to improve robustness against visually similar objects.
These methods represent the current state of the art on VOS benchmarks and are the most directly applicable general-purpose tools for our setting.
Nevertheless, all methods in this family share a fundamental architectural limitation for multi-object scenarios: each object is assigned an independent memory bank, and mask prediction, memory updates, and distractor management are performed per-object with no cross-object interaction.
There is no joint assignment step that enforces global identity consistency across all tracked objects simultaneously, making identity swaps between visually similar objects an inherent failure mode.
Additionally, these methods require manual mask or bounding-box initialization on the first frame of each target, limiting their applicability in autonomous deployment scenarios.
A related method, 3AM~\cite{sun20263am}, augments VOS with 3D-aware features from MASt3R~\cite{leroy2024muster}, but its code was not publicly available at the time of writing.
Our approach directly addresses these gaps through joint Hungarian assignment over all objects, explicit neighbor-context reasoning for disambiguation, and fully automatic detection-driven initialization.

\subsection{Online 3D Instance Segmentation in Indoor Scenes}

A separate line of work targets instance segmentation in indoor 3D scenes, effectively performing instance tracking as a camera traverses a static environment.
Methods such as AutoSeg3D~\cite{wang2026online}, ESAM~\cite{esam}, SAM2Object~\cite{sam2object}, and Any3DIS~\cite{any3dis} leverage RGB-D sensing and 3D geometric priors to associate 2D detections across frames, reporting performance on ScanNet~\cite{dai2017scannet} and ScanNet++~\cite{yeshwanth2023scannetpp} in terms of 3D instance segmentation metrics such as average precision.
While these methods achieve impressive results on their intended benchmarks, they almost universally require depth input and rely on 3D reconstruction pipelines that are unavailable in real-time monocular settings.
Furthermore, they report 3D metrics that are not directly comparable to the tracking metrics, needed to evaluate identity consistency, and no standardised MOT evaluation protocol currently exists for ScanNet or ScanNet++.
Our approach operates exclusively from monocular RGB, requires no depth sensing or 3D reconstruction, and is evaluated with tracking metrics that directly measure identity persistence, making it a practically deployable and directly comparable approach for the indoor navigation setting.


\section{REMIND Methodology}

\subsection{System Overview}

REMIND is an online multi-object re-identification tracker that operates frame-by-frame without future-frame lookahead. Given a sequence of RGB frames equipped with per-object segmentation masks produced by an off-the-shelf detector, the system assigns persistent identity labels to each detected object and maintains consistency across the entire sequence, including under long temporal gaps, occlusions, re-appearances, and visually similar instances.

\begin{figure*}[t]
  \centering
  \includegraphics[width=\textwidth]{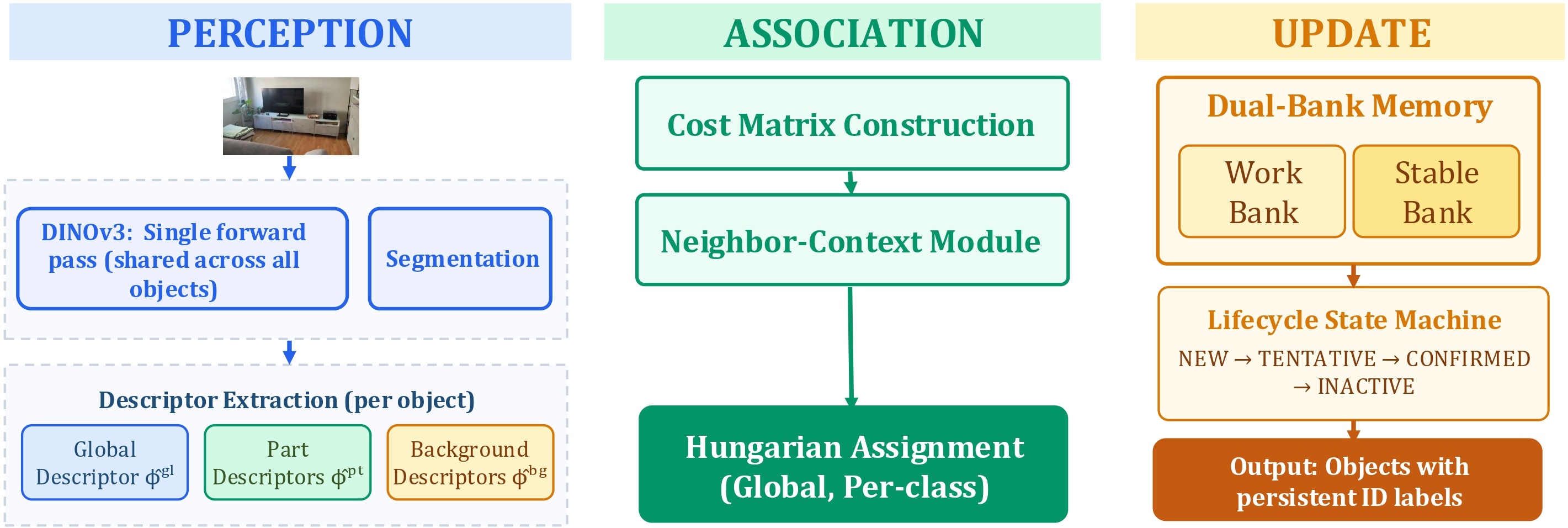}
  \caption{End-to-end REMIND pipeline. \textbf{Perception:} a single
  forward pass of a frozen DINOv3 ViT-S/16 produces patch features shared
  across all objects in the frame, while a YOLO segmentation detector supplies
  per-object masks; for each detection, the system extracts complementary
  global ($\hat{\boldsymbol{\phi}}^{\mathrm{gl}}$), part
  ($\hat{\boldsymbol{\phi}}^{\mathrm{pt}}$), and background
  ($\hat{\boldsymbol{\phi}}^{\mathrm{bg}}$) descriptors.
  \textbf{Association:} a multi-channel cost matrix is enriched with
  neighbour-context evidence before the Hungarian algorithm solves the
  bipartite assignment globally and per class. \textbf{Update:} matched
  detections feed a dual-bank (work/stable) memory and advance each track
  through the
  \textsc{New}$\to$\textsc{Tentative}$\to$\textsc{Confirmed}$\to$\textsc{Inactive}
  lifecycle, producing a frame annotated with persistent identity labels.}
  \label{fig:pipeline}
\end{figure*}

The pipeline follows a three-stage architecture executed sequentially for every incoming frame: (i)~\emph{Perception} -- extract visual descriptors from each detection; (ii)~\emph{Association} -- match current detections to previously tracked identities; (iii)~\emph{Update} -- incorporate new observations into persistent memory. No domain-specific training is required, and no manual mask or bounding-box prompts are needed at inference time; the only external requirement is an off-the-shelf segmentation detector with its own (pre-trained) class vocabulary, so the system is fully automatic for the deployment use-case but not zero-shot in the prompt-driven sense. Fig.~\ref{fig:pipeline} summarises the end-to-end architecture.

\begin{figure}[t]
  \centering
  \includegraphics[width=\linewidth]{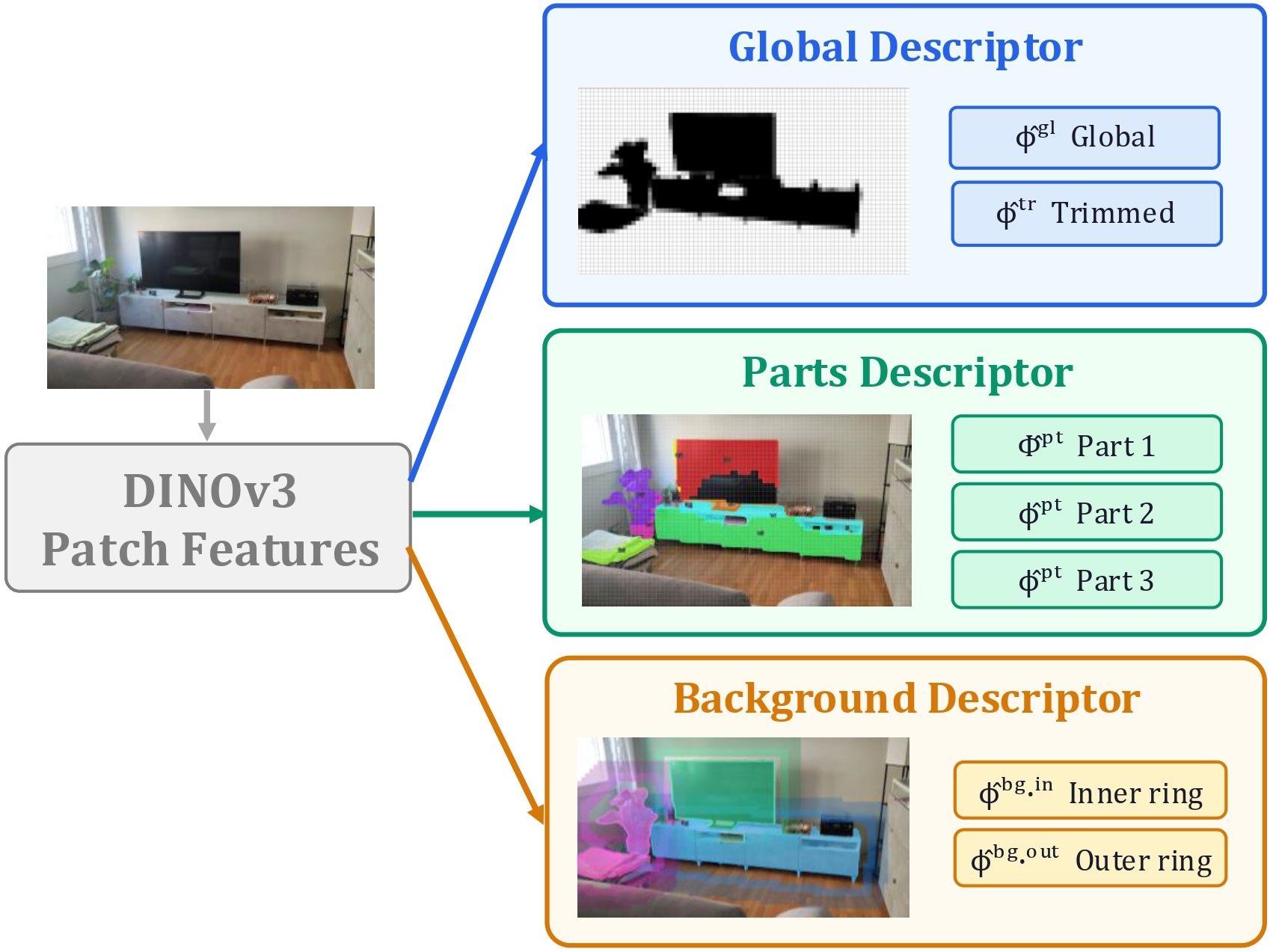}
  \caption{Three-channel descriptor extraction from a single shared DINOv3
  patch feature map. The \emph{global descriptor} pools the object as a whole
  into a coverage-weighted global descriptor
  $\hat{\boldsymbol{\phi}}^{\mathrm{gl}}$ and a boundary-robust trimmed-mean
  variant $\hat{\boldsymbol{\phi}}^{\mathrm{tr}}$. The \emph{parts descriptor}
  decomposes the object into $K$ semantic sub-regions, yielding
  multi-prototype part descriptors $\hat{\boldsymbol{\phi}}^{\mathrm{pt}}$. The
  \emph{background descriptor} samples local context through concentric inner
  ($\hat{\boldsymbol{\phi}}^{\mathrm{bg,in}}$) and outer
  ($\hat{\boldsymbol{\phi}}^{\mathrm{bg,out}}$) rings around the object. The
  three channels jointly feed the multi-channel similarity report used
  during association.}
  \label{fig:descriptors}
\end{figure}

\subsection{Perception and Descriptor Extraction}

The input frame $I_t \in \mathbb{R}^{H \times W \times 3}$ is resized to a target width while preserving its aspect ratio, then cropped so that both spatial dimensions are exact multiples of the Vision Transformer (ViT)~\cite{simeoni2025dinov3} patch size $p$ (typically $p=16$), yielding an aligned frame $I_a \in \mathbb{R}^{H_a \times W_a \times 3}$ with $H_a = p \lfloor H'/p \rfloor$ and $W_a = p \lfloor W'/p \rfloor$. A segmentation backend then produces a set of detections $\mathcal{D}_t = \{d_1,\ldots,d_N\}$, where each detection carries a binary pixel mask $M_d \in \{0,1\}^{H_a \times W_a}$, a class label $c_d \in \mathcal{C}$, and geometric attributes (bounding box, centroid, area).

A single forward pass of a frozen DINOv3~\cite{simeoni2025dinov3} ViT-S/16 on the aligned frame produces a dense patch-level feature map $F \in \mathbb{R}^{H_p \times W_p \times D}$, where $H_p = H_a/p$, $W_p = W_a/p$, and $D$ is the embedding dimension. We adopt DINOv3 as it is the most recent release in the self-supervised DINO family at the time of writing; the pipeline only relies on a frozen patch-level encoder and is therefore fully compatible with previous DINO and DINOv2 backbones, which can be substituted as drop-in replacements without architectural changes. Crucially, this single pass serves \emph{all} tracked objects in the scene simultaneously. For each detection $d$, a patch-level coverage map quantifies how much of each $p \times p$ patch cell falls within the instance mask:
\begin{equation}
  \mathrm{cov}(i,j) = \frac{1}{p^2} \sum_{(u,v)\in\mathrm{cell}(i,j)} M_d(u,v).
\end{equation}
A patch $(i,j)$ is considered valid for object $d$ if $\mathrm{cov}(i,j)>0$; coverage values further serve as spatial weights for descriptor pooling.

\subsubsection{Object Descriptors}

Two complementary representations are computed per detection (Fig.~\ref{fig:descriptors}). All $\ell_2$-normalized descriptors follow the unified notation $\hat{\boldsymbol{\phi}}^{(\cdot)}_d$, with a type superscript distinguishing each channel.

\emph{Global Descriptor} $\hat{\boldsymbol{\phi}}^{\mathrm{gl}}_d$. A coverage-weighted mean of valid patch features, $\ell_2$-normalized:
\begin{equation}
  \hat{\boldsymbol{\phi}}^{\mathrm{gl}}_d = \frac{\mathbf{g}_d}{\|\mathbf{g}_d\|_2}, \quad
  \mathbf{g}_d = \frac{\sum_{(i,j)\in\mathcal{P}_d} \mathrm{cov}(i,j)\,F(i,j)}{\sum_{(i,j)\in\mathcal{P}_d} \mathrm{cov}(i,j)},
\end{equation}
where $\mathcal{P}_d$ is the set of valid patches for detection $d$.

\emph{Trimmed-mean descriptor} $\hat{\boldsymbol{\phi}}^{\mathrm{tr}}_d$. A two-pass variant that first computes $\mathbf{g}_0$ as a preliminary mean, then retains only the top fraction $\rho$ of patches ranked by cosine similarity to $\mathbf{g}_0$ before re-pooling. This discards boundary patches whose features are contaminated by background content, which is particularly useful for small or irregular masks.

\subsubsection{Part Descriptors}

Part descriptors $\hat{\boldsymbol{\phi}}^{\mathrm{pt}}_{d,k}$ decompose each object into a set of semantic sub-regions, providing a multi-prototype representation that is more discriminative for objects with rich internal structure (e.g.\ chairs, backpacks, monitors). Two complementary methods are available.

\emph{K-means parts.} The $\ell_2$-normalized patch features of the object are clustered via K-means into $K$ groups (typically $K\in[3,5]$). Each cluster yields a part descriptor $\hat{\boldsymbol{\phi}}^{\mathrm{pt}}_{d,k}$ by weighted pooling of its member patches. Clusters below a minimum patch count are discarded, and a greedy merge step fuses clusters whose centroids have cosine similarity above a threshold $\tau_{\mathrm{merge}}$.

\emph{Attention parts.} Seed patches are selected by ranking object patches according to their in-degree in the multi-head self-attention graph, summed across layers. For each seed, a region is formed by taking the top-$k$ most-attended object patches, and the part descriptor $\hat{\boldsymbol{\phi}}^{\mathrm{pt}}_{d,k}$ is the weighted pool of that region. Each descriptor is $\ell_2$-normalized and accompanied by statistics including patch count, coverage support, and intra-cluster coherence.

\subsubsection{Background Descriptors}

Local background context is captured through concentric rings in patch space around each object. An inner ring of radius $r_{\mathrm{in}}$ patches and an outer ring of radius $r_{\mathrm{out}}>r_{\mathrm{in}}$ are constructed by dilating the object patch mask and subtracting the object region (plus a border-exclusion zone). Both radii adapt upward if the ring contains fewer patches than a minimum threshold.

For each ring, a global ring descriptor is computed as the weighted mean of patch features in the ring (with weights $1-\mathrm{cov}_\mathrm{obj}$ to downweight patches shared with the object), $\ell_2$-normalized; ring prototypes are additionally extracted via K-means clustering followed by inter-cluster merging. The inner- and outer-ring descriptors, denoted $\hat{\boldsymbol{\phi}}^{\mathrm{bg,in}}_d$ and $\hat{\boldsymbol{\phi}}^{\mathrm{bg,out}}_d$ respectively, are linearly blended with configurable weights into a combined background descriptor $\hat{\boldsymbol{\phi}}^{\mathrm{bg}}_d$.

At the end of the perception stage, each detection $d$ yields up to three descriptor channels stored in the matched identity's memory: (i)~a \emph{global channel} comprising $\hat{\boldsymbol{\phi}}^{\mathrm{gl}}_d$ and $\hat{\boldsymbol{\phi}}^{\mathrm{tr}}_d$; (ii)~a \emph{parts channel} containing $K$ part descriptors $\{\hat{\boldsymbol{\phi}}^{\mathrm{pt}}_{d,k}\}$ per active part method (K-means and/or attention); and (iii)~a \emph{background channel} with $\hat{\boldsymbol{\phi}}^{\mathrm{bg,in}}_d$, $\hat{\boldsymbol{\phi}}^{\mathrm{bg,out}}_d$, $\hat{\boldsymbol{\phi}}^{\mathrm{bg}}_d$, and ring prototypes. Per-patch feature vectors $\{\hat{F}(i,j)\}$ are used transiently during descriptor computation but are not persisted in memory.

\subsection{Object Memory Model}

Each tracked identity $o$ maintains a persistent appearance model structured around a dual-bank prototype memory, with independent banks maintained per descriptor type.

\subsubsection{Prototype Banks}

Each descriptor type, namely global ($\hat{\boldsymbol{\phi}}^{\mathrm{gl}}_d$ and $\hat{\boldsymbol{\phi}}^{\mathrm{tr}}_d$), part ($\{\hat{\boldsymbol{\phi}}^{\mathrm{pt}}_{d,k}\}$), and background ($\hat{\boldsymbol{\phi}}^{\mathrm{bg}}_d$), maintains two banks: a \emph{work bank} of recently observed embeddings that is actively updated every matched frame, and a \emph{stable bank} of consolidated embeddings promoted from the work bank once sufficient observations have accumulated. The stable bank provides a conservative, high-confidence representation for matching, while the work bank captures recent appearance variation. The dual-bank model operates independently of lifecycle state: objects in any active state (\textsc{Tentative} or \textsc{Confirmed}) populate their work bank, and \textsc{Inactive} objects retain both banks intact to support re-identification after long absences.

\subsubsection{Descriptor Update Protocol}

When detection $d$ is matched to identity $o$, the observed descriptor $\hat{\boldsymbol{\phi}}^{(\cdot)}_d$ is integrated into the work bank via the following protocol. First, a \emph{duplicate check} computes $s_{\max} = \max_k \cos(\hat{\boldsymbol{\phi}}^{(\cdot)}_d, \hat{\mathbf{e}}_k^{(o)})$ against all existing work prototypes. If $s_{\max}>\tau_{\mathrm{dup}}$, the observation is treated as a duplicate of the closest prototype, and the prototype is updated via an exponential moving average (EMA):
\begin{equation}
  \hat{\mathbf{e}}_k \leftarrow \mathrm{normalize}\!\left((1-\alpha)\,\hat{\mathbf{e}}_k + \alpha\,\hat{\boldsymbol{\phi}}^{(\cdot)}_d\right),
\end{equation}
where $\alpha$ is gated by $s_{\max}$ (higher similarity yields a more aggressive update) and scaled by the assignment confidence. If $s_{\max}<\tau_{\mathrm{dup}}$, a new prototype is inserted; if the bank is full, the most redundant or least-recently-used prototype is evicted. After insertion, any two work prototypes whose cosine similarity exceeds a merge threshold $\tau_{\mathrm{merge,internal}}$ are fused by weighted average. Work prototypes that accumulate sufficient observations are promoted to the stable bank. During long object absences, the memory is retained intact (unlike MOT trackers, which kill tracks at scene exits), enabling re-identification after gaps of hundreds of frames.

\subsubsection{Lifecycle Management}

Each tracked object follows a state machine: 

\begin{equation}
\textsc{New}\to\textsc{Tentative}\to\textsc{Confirmed}\to\textsc{Inactive}. 
\end{equation}

Each matched frame increments the hit counter and resets the miss counter; unmatched frames increment misses. After $h_{\mathrm{confirm}}$ cumulative hits (Table~\ref{tab:thresholds}), a \textsc{Tentative} object transitions to \textsc{Confirmed}. A \textsc{Confirmed} object exceeding $m_{\mathrm{max}}$ consecutive misses becomes \textsc{Inactive} but is not removed from memory, preserving its complete appearance model (both work and stable banks) for future re-identification. \textsc{Tentative} objects exceeding their miss budget and \textsc{Inactive} objects exceeding a time-to-live ($\mathrm{TTL}$; Table~\ref{tab:thresholds}) are eventually removed from memory.

\subsection{Contextual Reasoning}

A key architectural feature of REMIND is its ability to leverage spatial co-occurrence relationships between objects, serving two complementary roles. First, it disambiguates visually similar candidates by incorporating neighborhood evidence into the association cost matrix. Second, it provides supporting evidence that can rescue matches that are too weak to be resolved on appearance alone, for instance when an object is partially occluded or viewed from an unusual angle. Additionally, when contextual evidence strongly contradicts a candidate identity, a \emph{context veto} mechanism suppresses that candidate entirely, preventing erroneous assignments driven by misleading appearance similarity. These capabilities are not implemented by per-object-independent VOS methods.

\subsubsection{Co-occurrence Neighbor Graph}

For each tracked object $o$, historical neighbor co-occurrence frequencies are maintained as an exponentially-weighted probability kernel over co-visible object IDs. This graph records which other objects tend to appear together with $o$ across frames, building a stable spatial neighborhood model over time.

\subsubsection{Spatial Distance Neighbor Graph}

Pairwise spatial relations (normalized center distance, contact, containment, relative scale) between co-visible objects $o$ and $o'$ are accumulated per directed edge $(o, o')$ as a statistical summary. This distance graph enables the system to triangulate object identities from their spatial arrangement when appearance alone is ambiguous, which is a critical capability for scenes with multiple instances of the same object class (e.g.\ two identical chairs side by side).

\subsubsection{Neighbor-Sets Context Layer}

When enabled, the neighbor-sets module enriches the association cost matrix using contextual evidence. Given the current frame's reliably matched objects (anchors), a neighborhood hypothesis is generated specifying which tracked objects $o$ are expected to be present based on historical co-occurrence. Each detection-to-object candidate pair $(d, o)$ receives a bounded contextual bonus or penalty: candidates compatible with the expected neighborhood receive a positive adjustment (capped at $\delta_+$; Table~\ref{tab:thresholds}), while candidates clearly incompatible receive a negative adjustment (capped at $\delta_-$; Table~\ref{tab:thresholds}). The influence magnitude scales with a quality metric that combines neighborhood model coverage, maturity, density, and pruning effectiveness.

\subsection{Association}

\subsubsection{Multi-Channel Similarity Reports}

For each detection $d$ and each tracked object $o$ of the same class, a \emph{similarity report} aggregates evidence from three channels. The object channel scores $\hat{\boldsymbol{\phi}}^{\mathrm{gl}}_d$ against stored appearance prototypes:
\begin{equation}
  s_{\mathrm{obj}}(d,o) = \max_k \cos\!\left(\hat{\boldsymbol{\phi}}^{\mathrm{gl}}_d,\; \hat{\mathbf{e}}_k^{(o)}\right).
\end{equation}
The parts channel computes, for each active part method, the mean cosine similarity of the top-$k$ best-matched part pairs:
\begin{equation}
  s_{\mathrm{parts}}(d,o) = \frac{1}{\min(k,|\mathcal{P}_d|)} \sum_{i=1}^{\min(k,|\mathcal{P}_d|)} \max_j \cos\!\left(\hat{\boldsymbol{\phi}}^{\mathrm{pt}}_{d,i},\; \hat{\boldsymbol{\phi}}^{\mathrm{pt}}_{o,j}\right).
\end{equation}
The background channel compares observed ring descriptors against the stored background model prototypes using the same best-match scheme, with inner and outer terms combined by configurable weights.

\subsubsection{Quality-Weighted Score Combination}

The three channel scores are combined into a single similarity score $s_{\mathrm{sim}}(d,o)$ via a quality-aware weighted sum. Each channel $c\in\{\mathrm{obj},\mathrm{bg},\mathrm{parts}\}$ carries a nominal weight $w_c$ and a quality factor $q_c\in[0,1]$ derived from the detection's feature richness (valid patch count, part count, ring sizes, mask quality):
\begin{equation}
  s_{\mathrm{sim}}(d,o) = \frac{\sum_c w_c\,q_c^{\mathrm{eff}}\,s_c(d,o)}{\sum_c w_c\,q_c^{\mathrm{eff}}},
\end{equation}
where the denominator re-normalizes when channels are unavailable. A floor mechanism ensures that no channel's effective weight collapses to zero even for low-quality detections.

\subsubsection{Ambiguity Diagnosis and Reliable Anchors}

Each similarity report is classified as \textsc{Strong} (best candidate has high score and a clear margin $\delta_{\mathrm{confirm}}$ over the second-best), \textsc{Ambiguous} (two or more candidates have similar scores), or \textsc{Weak} (no candidate exceeds the match threshold $\tau_{\mathrm{match}}$). Before global assignment, the system identifies \emph{reliable anchor pairs} -- detections that are unambiguously \textsc{Strong} -- which serve as stable reference points for the contextual and disambiguation layers.

\subsubsection{Global Assignment}

The assignment problem is solved globally per semantic class using the Hungarian algorithm on a bipartite cost matrix. High-confidence pairs with score $\geq \tau_{\mathrm{lock}}$ and a sufficient gap are first \emph{locked} (pre-assigned and removed from the cost matrix) to prevent the global optimization from breaking obvious matches. Virtual ``new object'' columns are added to the matrix so that detections with no sufficiently similar existing track are assigned to create a new identity; the dummy score adapts per detection via a confidence-aware formula.

\subsubsection{Post-Assignment Guards}

Several mechanisms refine the raw Hungarian output. Detections assigned to a match but with \textsc{Ambiguous} similarity are flagged as \emph{AmbiguousTracks} -- temporary entities that accumulate observations across frames until the ambiguity resolves or a $\mathrm{TTL}$ expires (Table~\ref{tab:thresholds}). Detections assigned as ``new'' but with moderate similarity to existing objects become \emph{ProvisionalNewTracks} rather than immediately confirmed identities, preventing premature proliferation when a re-appearing object is temporarily hard to match.

When a group of detections maps ambiguously onto a known closed set of object IDs, a relational disambiguator resolves the assignment by comparing observed pairwise spatial relations (center distances, containment, contact) of the current detections against the historical distance graph of each candidate identity pair. Anchor-based scoring additionally leverages reliably matched objects as reference points to triangulate the correct assignment through a combined visual-plus-relational score.

\subsection{Memory Update}

Once associations are decided, the update stage incorporates new observations into persistent memory using the dual-bank EMA protocol described in Section~III-C. Update intensity is gated by match quality: \textsc{Strong} matches trigger full updates (insert, merge, promote, EMA); \textsc{Ambiguous} matches apply a safe-mode EMA-only update with a reduced $\alpha$ scale; \textsc{Weak} matches leave the memory unchanged. This gating prevents unreliable associations from corrupting the stored appearance model.

Part and background prototype banks are updated analogously to the global appearance channel, following the same insert-or-EMA policy. The co-occurrence neighbor graph and distance neighbor graph are updated each frame based on the confirmed co-visible identity set, accumulating episode counts and exponentially-weighted frequency kernels per edge.

\section{Results and Discussion}

\subsection{Datasets}

We evaluate REMIND on two complementary indoor benchmarks: a purpose-built custom video and the public dataset ScanNet++~\cite{yeshwanth2023scannetpp}.

\paragraph{Custom video} The custom benchmark is a single monocular sequence of 2\,min~17\,s recorded specifically for the application under study, in which we attempted to condense the main difficulties of long-term indoor re-identification into one continuous trajectory. The camera starts inside a room, transitions to a second, more cluttered room, then leaves both spaces, undergoes a strong drop in illumination, and finally re-enters the original room. The scene is intentionally populated with multiple visually similar instances of the same category, such as laptops, remote controls, forks, and scissors, in order to stress same-class disambiguation, appearance drift, and identity recovery after extended occlusions and revisits. After an extensive review of publicly available resources, we could not find a dataset that jointly reproduces these conditions for indoor navigation, namely controlled revisits, severe illumination changes, and dense same-class clutter under a single continuous monocular sequence. As part of future work, we plan to extend this benchmark into a broader dataset for the research community; the annotated \href{https://cvarbananas.myqnapcloud.com/share.cgi?ssid=571aa7a9fcb942c2af0e1ddbb71a6c4a}{custom video dataset} is already open-sourced.

\paragraph{ScanNet++} ScanNet++~\cite{yeshwanth2023scannetpp} is a high-fidelity dataset of indoor 3D scenes captured with a sub-millimetre laser scanner, registered 33\,megapixel DSLR images, and commodity RGB-D streams from a mobile device, comprising over one thousand scenes annotated with an open vocabulary of semantic and instance labels. From the released 3D annotations, we rasterised the trajectories of all selected sequences and projected the annotated objects into 2D image space, yielding per-frame instance masks consistent with the original 3D identities. We adopt the 83-class vocabulary defined by the dataset's own ``bench'' split, in contrast to the larger ``raw'' vocabulary, as it already excludes categories that are not relevant for indoor navigation, such as floors, walls, and ceilings. ScanNet++ is, to the best of our knowledge, the most suitable public dataset for this study, as it combines indoor navigation trajectories, dense instance-level annotation, and a meaningful proportion of visually similar objects. Nevertheless, it does not fully represent the target application: a non-negligible fraction of the annotated objects are very small fixtures with limited relevance for navigation, and most sequences remain confined to a single physical space, which limits the ability to evaluate long-term behaviour beyond the extent of an individual sequence.

\subsection{Results}

We benchmark REMIND against DAM4SAM~\cite{videnovic2025CVPR}, which constitutes the most directly applicable general-purpose tracker for our setting. DAM4SAM represents the current state of the art on VOS benchmarks, extends the SAM2 memory architecture to handle object disappearance and reappearance over long sequences, and introduces distractor-aware memory updates that suppress contamination from visually similar objects. The remaining paradigms reviewed previously, including multi-object tracking, person and vehicle re-identification, frame-to-frame generic association, and 3D-aware indoor instance segmentation, cover only subsets of the requirements imposed by long-term multi-object re-identification of generic indoor objects from monocular RGB. DAM4SAM is therefore selected as the natural and most demanding baseline. We additionally report MASA~\cite{li2024masa}, a strong tracking-by-detection method that learns generic instance association from segmentation proposals and can operate on the same per-frame masks or detections used by REMIND.

Two evaluation regimes are reported for each benchmark. In the first, the trackers consume ground-truth masks as input, which isolates the identity association and memory components from any error introduced by segmentation, and is therefore essential to characterise the intrinsic re-identification behaviour of REMIND decoupled from detector quality. In the second, masks are obtained from a custom YOLO11-Large segmentation detector~\cite{khanam2024yolov11}, chosen for its versatility across object categories and its real-time inference capabilities, which exercises the full end-to-end pipeline under realistic operating conditions. The detector is used differently by the systems, as imposed by their respective architectures. Under the GT-mask regime, REMIND and MASA consume a ground-truth mask for every annotated observation at every frame, while DAM4SAM is given the ground-truth mask of an object only the first time its identity is observed in the sequence and from then on propagates it internally through its own memory model; this asymmetry is not a design choice but an intrinsic property of DAM4SAM's algorithm, which is constructed around a one-shot initialisation followed by per-object memory propagation, and supplying a fresh GT mask on every frame would replace its core mechanism rather than evaluate it. The GT-mask regime should therefore be read as the configuration that is closest to DAM4SAM's native operating mode (one-shot initialisation from a privileged mask, then propagation) and that simultaneously preserves the standard tracking setting for REMIND and MASA. Under the YOLO regime, REMIND and MASA are fed YOLO detections at every frame, and any DAM4SAM detection that cannot be matched to an existing track instantiates a new memory bank. We adopt this configuration because it is the least invasive way of evaluating DAM4SAM end-to-end without leaking privileged ground truth at test time: a one-shot YOLO initialisation followed by silent propagation would instead tie its recovery capability to whichever frame the detector first fires on. Only the minimal coordination logic required for this purpose is added on top of DAM4SAM. We acknowledge that this stresses DAM4SAM beyond its intended one-shot paradigm, and that part of the 66.9\% out-of-memory failure rate reported below stems from every unmatched detection instantiating a new SAM2 memory bank; the GT-mask regime is reported precisely to disentangle this adaptation cost from the underlying re-identification capability.

For MASA, we use the official MASA-GDINO configuration as the base model, but adapt the evaluation setup to the long-term object re-identification setting considered in this work. Detections are processed in a class-aware manner, so candidate associations are restricted to instances of the same semantic class, matching the category-aware comparison used by REMIND. Spatial distance filtering is disabled because the target behaviour involves large camera motion, viewpoint changes, disappearances, and long temporal gaps, where image-space proximity is not a reliable identity cue. MASA memory is kept active over the full scene rather than limited to short temporal windows, favouring persistent identity recovery after revisits, and the embedding momentum is set to 0.3, which gave the best empirical behaviour among the tested values. These modifications are not part of the default MASA configuration; they were introduced to make the comparison more favourable to MASA and better aligned with persistent indoor re-identification.

Beyond the standard MOT and HOTA metrics, two re-identification-specific scores are reported. For each ground-truth identity, the per-frame timeline is segmented into maximal continuous visibility intervals, and the predicted label most consistent with that identity is taken as its reference. Every interval after the first counts as a \emph{reappearance opportunity}: the \emph{Recovery} rate is the fraction of such intervals in which the reference label is matched on at least one of their frames, while the \emph{Hard recovery} rate is the stricter fraction in which the label is recovered already on the very first frame of the new interval, with no transient identity drift. For instance, given a GT object with visibility intervals $[1,4]$, $[8,9]$, $[14,15]$, $[17,18]$ and predicted labels $[5,5,5,5]$, $[6,5]$, $[5,5]$, $[6,6]$, there are three reappearance opportunities and two correct recoveries, one of which is hard (interval $[14,15]$, recovered from its first frame) and one permissive (interval $[8,9]$, recovered only from the second frame).

\paragraph{Custom video}
Tables~\ref{tab:custom_video_results} and~\ref{tab:custom_video_hota} report the main results on the custom video benchmark. This is the setting where REMIND delivers its largest gains. Under ground-truth masks, IDF1 rises by nearly 20 points, from 70.53\% to 90.35\%, with AssA increasing from 77.59\% to 94.01\% and IDR from 64.64\% to 90.35\%; recovery improves from 69.15\% to 85.46\% and the ID-switch rate drops from 9.27\% to 5.74\%. MASA remains substantially below both methods in this regime, with 53.84\% IDF1 and 45.74\% recovery---more than 36 IDF1 points behind REMIND---although it obtains the lowest switch rate. The end-to-end YOLO regime preserves the same ordering, with IDF1 rising from 64.54\% to 75.45\%, recovery from 65.96\% to 79.08\%, hard recovery from 39.72\% to 44.33\%, and AssA from 73.52\% to 94.03\%; MASA reaches 45.76\% IDF1 and 38.65\% recovery. DetA is reported as N/A for REMIND and MASA under the GT-mask regime because these systems, by construction, evaluate every annotated mask, so DetA collapses trivially to 100\% and carries no comparative information; for DAM4SAM, DetA remains meaningful because its propagated output does not cover all ground-truth observations. Under YOLO, where DetA becomes informative, MASA reports a near-complete 99.37\% against REMIND's 88.72\% despite both consuming identical detections: this gap is not a detector effect but a consequence of REMIND collapsing its association uncertainty into a single global identity assignment, so when several same-class detections compete for the same few objects the surplus ones are left without a confident identity and are effectively dropped, whereas MASA's per-detection matching retains them.


\begin{table}[H]
\centering
\scriptsize
\caption{Custom video results}
\label{tab:custom_video_results}

\begin{tabular*}{\columnwidth}{@{\extracolsep{\fill}}llcccc@{}}
\hline
\textbf{Method} & \textbf{Detector} & \textbf{IDF1} & \textbf{Recovery} & \textbf{Hard rec.} & \textbf{IDSW rate} \\
\hline

DAM4SAM &
\begin{tabular}[c]{@{}c@{}}
GT masks \\
init
\end{tabular}
& 70.53\% & 69.15\% & 46.81\% & 9.27\% \\

MASA &
\begin{tabular}[c]{@{}c@{}}
GT masks
\end{tabular}
& 53.84\% & 45.74\% & 37.94\% & \textbf{5.30\%} \\

REMIND &
\begin{tabular}[c]{@{}c@{}}
GT masks
\end{tabular}
& \textbf{90.35\%} & \textbf{85.46\%} & \textbf{58.16\%} & 5.74\% \\

\hline

DAM4SAM &
\begin{tabular}[c]{@{}c@{}}
YOLO init
\end{tabular}
& 64.54\% & 65.96\% & 39.72\% & 10.61\% \\

MASA &
\begin{tabular}[c]{@{}c@{}}
YOLO
\end{tabular}
& 45.76\% & 38.65\% & 34.04\% & \textbf{6.41\%} \\

REMIND &
\begin{tabular}[c]{@{}c@{}}
YOLO
\end{tabular}
& \textbf{75.45\%} & \textbf{79.08\%} & \textbf{44.33\%} & 8.60\% \\

\hline

\end{tabular*}

\end{table}


\begin{table}[H]
\centering
\scriptsize
\caption{Custom video HOTA metrics}
\label{tab:custom_video_hota}

\begin{tabular*}{\columnwidth}{@{\extracolsep{\fill}}llcccc@{}}
\hline
\textbf{Method} & \textbf{Detector} & \textbf{DetA} & \textbf{AssA} & \textbf{IDP} & \textbf{IDR} \\
\hline

DAM4SAM &
\begin{tabular}[c]{@{}c@{}}
GT masks \\
init
\end{tabular}
& 83.32\% & 77.59\% & 77.59\% & 64.64\% \\

MASA &
\begin{tabular}[c]{@{}c@{}}
GT masks
\end{tabular}
& N/A & 53.84\% & 53.84\% & 53.84\% \\

REMIND &
\begin{tabular}[c]{@{}c@{}}
GT masks
\end{tabular}
& N/A & \textbf{94.01\%} & \textbf{90.35\%} & \textbf{90.35\%} \\

\hline

DAM4SAM &
\begin{tabular}[c]{@{}c@{}}
YOLO init
\end{tabular}
& 78.25\% & 73.52\% & 73.52\% & 57.53\% \\

MASA &
\begin{tabular}[c]{@{}c@{}}
YOLO
\end{tabular}
& \textbf{99.37\%} & 45.90\% & 45.90\% & 45.61\% \\

REMIND &
\begin{tabular}[c]{@{}c@{}}
YOLO
\end{tabular}
& 88.72\% & \textbf{94.03\%} & \textbf{76.76\%} & \textbf{74.19\%} \\

\hline

\end{tabular*}

\end{table}

\paragraph{ScanNet++}
Tables~\ref{tab:scannet_results} and~\ref{tab:scannet_hota} extend the evaluation to ScanNet++. On the DAM4SAM-comparable subset, REMIND wins IDF1 in both regimes (66.67\% vs.\ 60.71\% for DAM4SAM and 57.01\% for MASA under ground-truth masks; 61.30\% vs.\ 57.97\% and 60.38\% under YOLO), so MASA is competitive under YOLO but weaker under ground-truth masks. The recovery-oriented gains are largest under ground-truth masks, where REMIND lifts recovery to 58.36\% (from 41.65\% and 42.65\%), hard recovery to 42.62\% (from 27.07\% and 30.61\%), and IDR to 66.20\% (from 50.28\% and 57.01\%); under YOLO it additionally raises AssA to 81.68\% (from 62.70\% and 69.73\%). The trade-off is identity purity: REMIND admits more recoveries at a higher ID-switch rate under ground-truth masks (16.28\% vs.\ 4.68\% and 13.96\%), while the ordering reverses under YOLO (11.14\% vs.\ 20.33\% and 14.13\%), consistent with a recovery-oriented design that prefers a plausible re-identification over a conservative new-track decision when the visual evidence supports either.

The DAM4SAM-comparable subset reported in Tables~\ref{tab:scannet_results} and~\ref{tab:scannet_hota} excludes 2.2\% of scenes under ground-truth masks and 66.9\% under YOLO, all due to VRAM out-of-memory on the NVIDIA GeForce RTX 4090 (24\,GB VRAM) used in our experiments; the forced YOLO rows use the V2 subset. The dominance of the YOLO failure rate over the ground-truth-mask one is a direct consequence of DAM4SAM's per-object memory architecture: every unmatched YOLO detection, including transient false positives and re-detections that fail to match an existing track, instantiates a new SAM2 memory bank, so VRAM grows with the cumulative track count rather than with the true object count. REMIND and MASA complete 100\% of evaluated scenes in both regimes; the all-scenes rows in the same tables therefore constitute the more representative deployment figures. On the full ScanNet++ set, REMIND remains ahead of MASA under ground-truth masks (62.47\% vs.\ 54.89\% IDF1), whereas MASA attains the highest YOLO IDF1 (50.56\% vs.\ 48.07\%). This inversion reflects the benchmark more than the method: most ScanNet++ sequences stay confined to a single space with quasi-continuous visibility and few genuine long-gap revisits, reducing long-term re-identification to the frame-to-frame association that is precisely MASA's regime, while the persistent memory and neighbour context behind REMIND's 36-point margin on the custom video are structurally underused---the neighbour-context ablation costs 20.6 IDF1 points on the custom video but only 2.1 on ScanNet++, showing this machinery is nearly idle here rather than harmful. Even the YOLO edge is a detection-coverage effect rather than an association win: MASA wins DetA (69.44\% vs.\ 64.57\%) and hence IDF1, whereas REMIND holds clearly higher AssA (69.99\% vs.\ 61.69\%) and recovery (34.74\% vs.\ 34.10\%), and since both consume the same YOLO detections the DetA gap indicates REMIND reports fewer of them, consistent with its ambiguity-aware guards deferring uncertain observations and gating low-confidence tiny masks, which ScanNet++ contains in abundance. Closing this detection gap therefore remains the main avenue for further improvement.


\begin{table}[H]
\centering
\scriptsize
\caption{ScanNet++ results. ``Forced'' denotes REMIND/MASA evaluated on the DAM4SAM-comparable subset for direct comparison; forced YOLO uses the V2 subset.}
\label{tab:scannet_results}

\resizebox{\columnwidth}{!}{%
\begin{tabular}{@{}llccccc@{}}
\hline
\textbf{Method} & \textbf{Detector} & \textbf{IDF1} & \textbf{Recovery} & \textbf{Hard rec.} & \textbf{IDSW rate} &
\begin{tabular}[c]{@{}c@{}}\textbf{Failed} \\ \textbf{Scenes}\end{tabular} \\
\hline

DAM4SAM &
\begin{tabular}[c]{@{}c@{}}
GT masks \\
init
\end{tabular}
& 60.71\% & 41.65\% & 27.07\% & \textbf{4.68\%} & 2.2\% \\

MASA &
\begin{tabular}[c]{@{}c@{}}
GT masks
\end{tabular}
& 57.01\% & 42.65\% & 30.61\% & 13.96\% & \textbf{0\%} (forced 2.2\%) \\

REMIND &
\begin{tabular}[c]{@{}c@{}}
GT masks
\end{tabular}
& \textbf{66.67\%} & \textbf{58.36\%} & \textbf{42.62\%} & 16.28\% & \textbf{0\%} (forced 2.2\%) \\

\hline

DAM4SAM &
\begin{tabular}[c]{@{}c@{}}
YOLO init
\end{tabular}
& 57.97\% & \textbf{53.59\%} & \textbf{35.59\%} & 20.33\% & 66.9\% \\

MASA &
\begin{tabular}[c]{@{}c@{}}
YOLO
\end{tabular}
& 60.38\% & 46.89\% & 29.42\% & \textbf{10.79\%} & \textbf{0\%} (forced 66.9\%) \\

REMIND &
\begin{tabular}[c]{@{}c@{}}
YOLO
\end{tabular}
& \textbf{61.30\%} & 46.23\% & 27.21\% & 11.14\% & \textbf{0\%} (forced 66.9\%) \\

\hline

\begin{tabular}[c]{@{}l@{}}MASA \\ (All Scenes)\end{tabular} &
\begin{tabular}[c]{@{}c@{}}
GT masks
\end{tabular}
& 54.89\% & 40.60\% & 28.96\% & \textbf{14.12\%} & \textbf{0\%} \\

\begin{tabular}[c]{@{}l@{}}REMIND \\ (All Scenes)\end{tabular} &
\begin{tabular}[c]{@{}c@{}}
GT masks
\end{tabular}
& \textbf{62.47\%} & \textbf{55.12\%} & \textbf{39.53\%} & 19.17\% & \textbf{0\%} \\

\hline

\begin{tabular}[c]{@{}l@{}}MASA \\ (All Scenes)\end{tabular} &
\begin{tabular}[c]{@{}c@{}}
YOLO
\end{tabular}
& \textbf{50.56\%} & 34.10\% & \textbf{18.63\%} & \textbf{7.65\%} & \textbf{0\%} \\

\begin{tabular}[c]{@{}l@{}}REMIND \\ (All Scenes)\end{tabular} &
\begin{tabular}[c]{@{}c@{}}
YOLO
\end{tabular}
& 48.07\% & \textbf{34.74\%} & 16.27\% & 12.19\% & \textbf{0\%} \\

\hline
\end{tabular}%
}

\end{table}


\begin{table}[H]
\centering
\scriptsize
\caption{ScanNet++ HOTA metrics. ``Forced'' denotes REMIND/MASA evaluated on the DAM4SAM-comparable subset for direct comparison; forced YOLO uses the V2 subset.}
\label{tab:scannet_hota}

\resizebox{\columnwidth}{!}{%
\begin{tabular}{@{}llccccc@{}}
\hline
\textbf{Method} & \textbf{Detector} & \textbf{DetA} & \textbf{AssA} & \textbf{IDP} & \textbf{IDR} &
\begin{tabular}[c]{@{}c@{}}\textbf{Failed} \\ \textbf{Scenes}\end{tabular} \\
\hline

DAM4SAM &
\begin{tabular}[c]{@{}c@{}}
GT masks \\
init
\end{tabular}
& 59.37\% & \textbf{76.60\%} & \textbf{76.60\%} & 50.28\% & 2.2\% \\

MASA &
\begin{tabular}[c]{@{}c@{}}
GT masks
\end{tabular}
& N/A & 57.01\% & 57.01\% & 57.01\% & \textbf{0\%} (forced 2.2\%) \\

REMIND &
\begin{tabular}[c]{@{}c@{}}
GT masks
\end{tabular}
& N/A & 71.44\% & 67.14\% & \textbf{66.20\%} &  \textbf{0\%} (forced 2.2\%) \\

\hline

DAM4SAM &
\begin{tabular}[c]{@{}c@{}}
YOLO init
\end{tabular}
& \textbf{85.98\%} & 62.70\% & 62.70\% & 53.91\% & 66.9\% \\

MASA &
\begin{tabular}[c]{@{}c@{}}
YOLO
\end{tabular}
& 76.37\% & 69.73\% & 69.73\% & 53.25\% & \textbf{0\%} (forced 66.9\%) \\

REMIND &
\begin{tabular}[c]{@{}c@{}}
YOLO
\end{tabular}
& 71.88\% & \textbf{81.68\%} & \textbf{70.06\%} & \textbf{54.49\%} & \textbf{0\%} (forced 66.9\%) \\

\hline

\begin{tabular}[c]{@{}l@{}}MASA \\ (All Scenes)\end{tabular} &
\begin{tabular}[c]{@{}c@{}}
GT masks
\end{tabular}
& N/A & 54.89\% & 54.89\% & 54.89\% & \textbf{0\%} \\

\begin{tabular}[c]{@{}l@{}}REMIND \\ (All Scenes)\end{tabular} &
\begin{tabular}[c]{@{}c@{}}
GT masks
\end{tabular}
& N/A & \textbf{67.32\%} & \textbf{62.86\%} & \textbf{62.08\%} & \textbf{0\%} \\

\hline

\begin{tabular}[c]{@{}l@{}}MASA \\ (All Scenes)\end{tabular} &
\begin{tabular}[c]{@{}c@{}}
YOLO
\end{tabular}
& \textbf{69.44\%} & 61.69\% & \textbf{61.69\%} & \textbf{42.84\%} & \textbf{0\%} \\

\begin{tabular}[c]{@{}l@{}}REMIND \\ (All Scenes)\end{tabular} &
\begin{tabular}[c]{@{}c@{}}
YOLO
\end{tabular}
& 64.57\% & \textbf{69.99\%} & 57.94\% & 41.07\% & \textbf{0\%} \\

\hline
\end{tabular}%
}

\end{table}

\begin{figure}[H]
\centering

\begin{tikzpicture}
\begin{axis}[
    ybar,
    bar width=6pt,
    width=\columnwidth,
    height=5.5cm,
    ymin=0,
    ymax=100,
    ylabel={Rate (\%)},
    xlabel={Same-class distractors bin},
    symbolic x coords={0,1,2,3-4,5-8,9-16,17+},
    xtick=data,
    enlarge x limits=0.15,
    legend style={
        at={(0.5,1.18)},
        anchor=south,
        legend columns=3
    },
    title={IDF1 by same-class distractors},
    title style={yshift=-6pt}
]

\addplot coordinates {
(0,99.43)
(1,82.29)
(2,71.60)
(3-4,62.60)
(5-8,51.61)
(9-16,43.45)
(17+,29.84)
};

\addplot coordinates {
(0,72.13)
(1,64.67)
(2,61.31)
(3-4,57.56)
(5-8,53.45)
(9-16,47.45)
(17+,38.30)
};

\addplot coordinates {
(0,62.09)
(1,57.80)
(2,53.44)
(3-4,53.08)
(5-8,49.46)
(9-16,44.08)
(17+,39.22)
};

\legend{REMIND,DAM4SAM,MASA}

\end{axis}
\end{tikzpicture}

\caption{Behaviour analysis under different same-class distractor density conditions on ScanNet++ using ground-truth masks.}
\label{fig:idf1_same-class_distractors}

\end{figure}
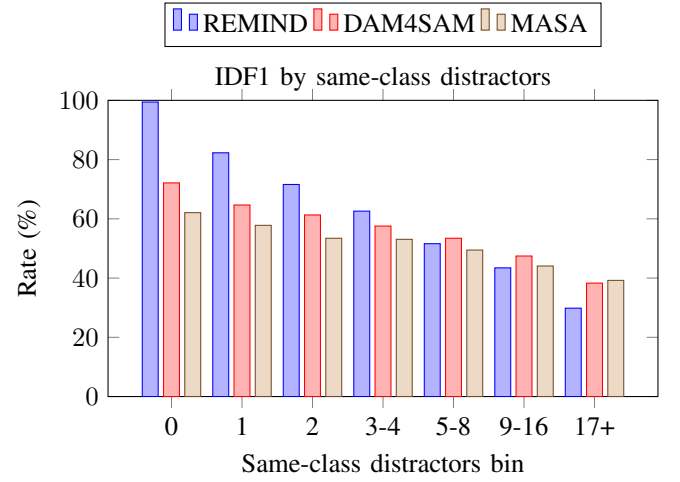

\paragraph{Behaviour analysis}
Figure~\ref{fig:idf1_same-class_distractors} isolates the main failure mode. REMIND strongly outperforms DAM4SAM and MASA when same-class distractors are few or moderate, where appearance, local background, and relational context provide enough evidence to disambiguate identities. As the count grows, the gap narrows and eventually inverts: MASA exhibits the flattest degradation of the three (62.09\% down to 39.22\%), its learned generic embeddings decaying gracefully with distractor density but starting far lower, so REMIND's advantage is front-loaded in precisely the low-to-moderate regime that dominates the target application. The crossover in the densest bins is consistent with the elevated ID-switch rate on ScanNet++: repeated instances of the same category remain the hardest case for the current system.


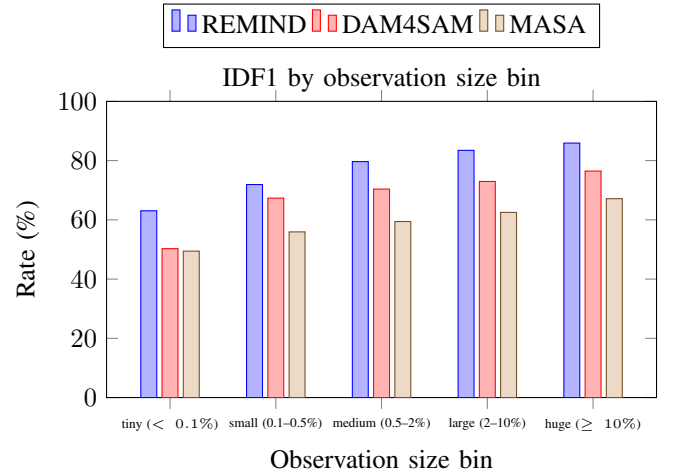
\begin{figure}[H]
\centering

\begin{tikzpicture}
\begin{axis}[
    ybar,
    bar width=6pt,
    width=\columnwidth,
    height=5.5cm,
    ymin=0,
    ymax=100,
    ylabel={Rate (\%)},
    xlabel={Observation size bin},
    symbolic x coords={tiny,small,medium,large,huge},
    xtick=data,
    xticklabels={
        tiny ($<0.1\%$),
        small (0.1--0.5\%),
        medium (0.5--2\%),
        large (2--10\%),
        huge ($\geq 10\%$)
    },
    xticklabel style={
        font=\tiny,
        align=center,
        text width=1.8cm
    },
    enlarge x limits=0.15,
    legend style={
        at={(0.5,1.18)},
        anchor=south,
        legend columns=3
    },
    title={IDF1 by observation size bin},
    title style={yshift=-6pt}
]

\addplot coordinates {
(tiny,63.06)
(small,71.93)
(medium,79.67)
(large,83.47)
(huge,85.92)
};

\addplot coordinates {
(tiny,50.27)
(small,67.34)
(medium,70.39)
(large,72.96)
(huge,76.47)
};

\addplot coordinates {
(tiny,49.44)
(small,55.94)
(medium,59.44)
(large,62.52)
(huge,67.15)
};

\legend{REMIND,DAM4SAM,MASA}

\end{axis}
\end{tikzpicture}

\caption{Behaviour analysis under different observation mask size conditions on ScanNet++ using ground-truth masks.}
\label{fig:idf1_observation_size}

\end{figure}

Figure~\ref{fig:idf1_observation_size} breaks down performance by observation size. REMIND improves IDF1 across all bins, including tiny and small masks. This supports the use of multiple descriptor channels: when the object mask is small and pure appearance becomes noisy, local background and part-level evidence provide complementary information. Larger objects are easier for all methods, including MASA, but REMIND retains a consistent advantage.


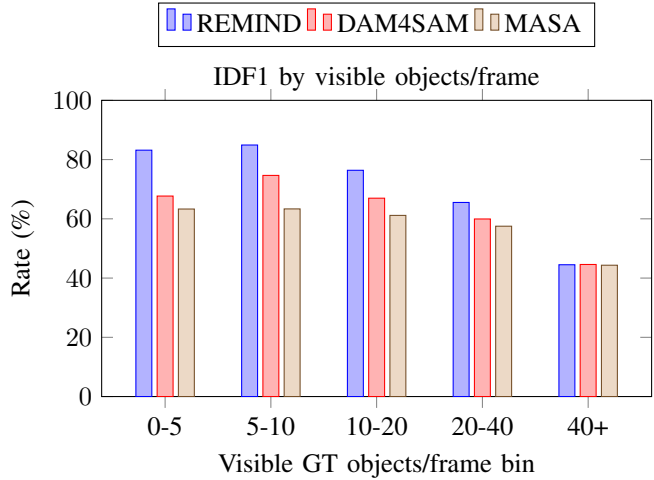
\begin{figure}[H]
\centering

\begin{tikzpicture}
\begin{axis}[
    ybar,
    bar width=6pt,
    width=\columnwidth,
    height=5.5cm,
    ymin=0,
    ymax=100,
    ylabel={Rate (\%)},
    xlabel={Visible GT objects/frame bin},
    symbolic x coords={0-5,5-10,10-20,20-40,40+},
    xtick=data,
    enlarge x limits=0.15,
    legend style={
        at={(0.5,1.18)},
        anchor=south,
        legend columns=3
    },
    title={IDF1 by visible objects/frame},
    title style={yshift=-6pt}
]

\addplot coordinates {
(0-5,83.18)
(5-10,84.92)
(10-20,76.39)
(20-40,65.53)
(40+,44.50)
};

\addplot coordinates {
(0-5,67.69)
(5-10,74.66)
(10-20,66.97)
(20-40,59.95)
(40+,44.59)
};

\addplot coordinates {
(0-5,63.31)
(5-10,63.35)
(10-20,61.16)
(20-40,57.51)
(40+,44.34)
};

\legend{REMIND,DAM4SAM,MASA}

\end{axis}
\end{tikzpicture}

\caption{Behaviour analysis under different visible object density conditions on ScanNet++ using ground-truth masks.}
\label{fig:idf1_visible_objects}

\end{figure}

Figure~\ref{fig:idf1_visible_objects} analyses scene complexity through per-frame visible-object density. REMIND performs best at low and moderate densities, while DAM4SAM and MASA follow the same downward trend as density grows. The gap narrows in crowded frames, showing that the bottleneck is not isolated object re-identification but global identity competition, where same-class candidates accumulate within a single global assignment step.


\paragraph{Runtime and memory}
Table~\ref{tab:efficiency_resources} summarises runtime and memory usage for the ground-truth-mask settings. REMIND is faster than DAM4SAM on both benchmarks: average loop time drops from 640.92\,ms to 342.61\,ms on the custom video and from 737.20\,ms to 500.83\,ms on ScanNet++, with the gain preserved at the 95th percentile. MASA is faster still in these logs (109.00\,ms on the custom video and 136.30\,ms on ScanNet++), although with lower recovery and association accuracy in the main metrics. Additionally, and purposely left out from this analysis, we experimentally verified that switching to YOLO further reduces REMIND's average loop on ScanNet++ to 348.08\,ms, not because the detector is cheaper than ground-truth masks but because YOLO produces fewer per-frame candidates, so the association and memory stages process less.

The memory profile is asymmetric. REMIND keeps peak allocated VRAM below 0.3\,GiB on both benchmarks, whereas DAM4SAM reaches 9.47\,GiB on the custom video and 21.54\,GiB on ScanNet++; MASA peaks at 1.67\,GiB and 1.70\,GiB, respectively. The trade-off is on the CPU side, where REMIND's peak RSS on ScanNet++ is 20.71\,GiB versus 6.75\,GiB for DAM4SAM and 7.19\,GiB for MASA. This is driven by per-frame scene density rather than total sequence length: loop time correlates more strongly with the mean number of visible objects per frame ($r{=}0.96$) and total per-frame distractors ($r{=}0.96$) than with total scene object count ($r{=}0.84$) or mean same-class count ($r{=}0.63$). In practice, the slowest cases are crowded frames with many simultaneous candidates. The 20.71\,GiB ScanNet++ figure should be read as an upper bound for an unusually dense benchmark in which every annotated fixture, including small items irrelevant for navigation, contributes a per-object dual prototype bank and a node in the co-occurrence and distance graphs. In the indoor-navigation regime for which REMIND is intended, where the robot tracks tens of navigation-relevant objects per scene rather than the hundreds of fixture-level annotations in ScanNet++, peak RSS stays well within mobile-platform budgets: on the custom video, which already contains a deliberately overpopulated mix of laptops, monitors, kitchen utensils, and other clutter, REMIND peaks at only 1.36\,GiB. The 20.71\,GiB figure is therefore representative of an exhaustive offline benchmark configuration rather than of on-robot deployment.


\begin{table}[H]
\centering
\scriptsize
\caption{Runtime and memory usage for the GT-mask settings used in the main comparison}
\label{tab:efficiency_resources}

\begin{tabular*}{\columnwidth}{@{\extracolsep{\fill}}llcccc@{}}
\hline
\textbf{Dataset} & \textbf{Method} & \textbf{Avg loop} & \textbf{P95 loop} & \textbf{Peak VRAM} & \textbf{Peak RSS} \\
 &  & \textbf{(ms)} & \textbf{(ms)} & \textbf{(GiB)} & \textbf{(GiB)} \\
\hline

Custom & DAM4SAM & 640.92 & 755.43 & 9.47 & 2.00 \\
Custom & MASA & \textbf{109.00} & \textbf{126.30} & 1.67 & 2.14 \\
Custom & REMIND & 342.61 & 488.59 & \textbf{0.16} & \textbf{1.36} \\

\hline

ScanNet++ & DAM4SAM & 737.20 & 1357.77 & 21.54 & \textbf{6.75} \\
ScanNet++ & MASA & \textbf{136.30} & \textbf{174.90} & 1.70 & 7.19 \\
ScanNet++ & REMIND & 500.83 & 1107.96 & \textbf{0.25} & 20.71 \\

\hline
\end{tabular*}

\end{table}

Taken together, REMIND is a recovery-oriented indoor re-identification tracker. Its largest gains arise under controlled revisits, severe illumination changes, and dense same-class clutter, where IDF1 improves by nearly 20 points over DAM4SAM and by more than 36 points over MASA under ground-truth masks. It also wins IDF1 in both forced regimes on ScanNet++ while completing every evaluated scene, including the 66.9\% of YOLO sequences on which DAM4SAM exhausts GPU memory. MASA becomes competitive only where the benchmark collapses toward frame-to-frame association, edging ahead in all-scenes YOLO IDF1 through detection coverage, but its lower recovery and AssA confirm that generic tracking-by-detection association does not fully solve long-term indoor identity recovery. The main residual limitation is identity purity in highly crowded same-class scenarios, where the system recovers more objects but admits more switches, and end-to-end performance remains sensitive to detector quality.

\paragraph{Diagnosing the next bottleneck} Beyond the collapsed metrics, REMIND's ambiguity handling exposes a useful diagnostic property of where to focus future improvements. On ScanNet++, 77.93\% of cases are firm decisions (81.12\% of them correct), 20.56\% are ambiguous candidate sets that contain the correct identity 56.86\% of the time but only collapse to it in 38.91\%, and the residual 1.51\% are \textsc{Weak} reports in which no candidate clears the match threshold and the detection is therefore deferred to a provisional or new track; on the custom benchmark, ambiguous sets contain the correct identity 98.29\% of the time and collapse correctly in 79.66\%. The gap between candidate-set hit rate and final collapse accuracy isolates the next actionable sub-problem, namely picking the correct identity out of a plausible set, and motivates strengthening the disambiguation stage rather than the recovery stage in future work.

\subsection{Ablation Study}

We ablate three design choices on both benchmarks: replacing the global Hungarian assignment with a greedy per-detection match (\texttt{greedy}), removing the local background channel (\texttt{no background}), and removing the neighbor-context layer (\texttt{no neighbors}). Results are reported in Tables~\ref{tab:ablation_results} and~\ref{tab:ablation_hota}.

The local background channel is the most critical component. Removing it collapses IDF1 from 90.35\% to 55.56\% on the custom video and from 62.47\% to 37.53\% on ScanNet++, with the ID-switch rate climbing to 51.73\% on ScanNet++. This confirms that local background ring descriptors carry most of the disambiguation signal between visually similar instances of the same category.

The neighbor-context layer is the second most impactful. Disabling it drops IDF1 from 90.35\% to 69.73\% on the custom video and from 62.47\% to 60.39\% on ScanNet++. The asymmetry is informative: neighbor reasoning is decisive under controlled revisits, where spatial co-occurrence patterns repeat reliably, but adds smaller residual gains on ScanNet++, whose shorter trajectories offer fewer opportunities to consolidate such patterns.

Replacing the global Hungarian assignment with a greedy per-detection match has a marginal effect on the custom video (90.35\% to 90.54\%, essentially noise) but degrades ScanNet++ both in IDF1 (62.47\% to 61.30\%) and in completion, with failed-scene rate rising from 0\% to 18.3\%. A comparable completion collapse is observed for the other two ablations, with failure rates of 22.6\% (no background) and 19.0\% (no neighbors). The failed scenes in all three ablations are caused by system RAM out-of-memory under a 24\,GB cap enforced for the ablation runs, in contrast to the main DAM4SAM failures, which are GPU VRAM out-of-memory on the same RTX 4090; in both cases the failure modes are driven by per-frame scene density rather than algorithmic divergence. Joint assignment, contextual reasoning, and the background channel therefore not only improve identity metrics but also prevent these catastrophic failures in dense or visually ambiguous scenes.


\begin{table}[H]
\centering
\scriptsize
\caption{Ablation study results}
\label{tab:ablation_results}

\resizebox{\columnwidth}{!}{%
\begin{tabular}{@{}llccccccc@{}}
\hline
\textbf{Method} & \textbf{Dataset} & \textbf{Ablation} & \textbf{IDF1} & \textbf{Recovery} & \textbf{Hard rec.} & \textbf{IDSW rate} &
\begin{tabular}[c]{@{}c@{}}\textbf{Failed} \\ \textbf{Scenes}\end{tabular} \\
\hline

REMIND & Custom Video & base           & 90.35\% & \textbf{85.46\%} & \textbf{58.16\%} & 5.74\%  & \textbf{0.0\%} \\
REMIND & Custom Video & greedy         & \textbf{90.54\%} & \textbf{85.46\%} & \textbf{58.16\%} & \textbf{5.71\%}  & \textbf{0.0\%} \\
REMIND & Custom Video & no background  & 55.56\% & 53.19\% & 27.66\% & 19.87\% & \textbf{0.0\%} \\
REMIND & Custom Video & no neighbors   & 69.73\% & 56.03\% & 28.72\% & 11.58\% & \textbf{0.0\%} \\

\hline

REMIND & ScanNet++ & base          & \textbf{62.47\%} & \textbf{55.12\%} & \textbf{39.53\%} & \textbf{19.17\%} & \textbf{0.0\%}  \\
REMIND & ScanNet++ & greedy        & 61.30\% & 53.85\% & 38.31\% & 19.74\% & 18.3\% \\
REMIND & ScanNet++ & no background & 37.53\% & 31.74\% & 17.54\% & 51.73\% & 22.6\% \\
REMIND & ScanNet++ & no neighbors  & 60.39\% & 51.71\% & 35.68\% & 20.26\% & 19.0\% \\

\hline
\end{tabular}%
}

\end{table}


\begin{table}[H]
\centering
\scriptsize
\caption{Ablation study HOTA metrics}
\label{tab:ablation_hota}

\resizebox{\columnwidth}{!}{%
\begin{tabular}{@{}llccccccc@{}}
\hline
\textbf{Method} & \textbf{Dataset} & \textbf{Ablation} & \textbf{DetA} & \textbf{AssA} & \textbf{IDP} & \textbf{IDR} &
\begin{tabular}[c]{@{}c@{}}\textbf{Failed} \\ \textbf{Scenes}\end{tabular} \\
\hline

REMIND & Custom Video & base           & N/A & 94.01\% & 90.35\% & 90.35\% & \textbf{0.0\%} \\
REMIND & Custom Video & greedy         & N/A & \textbf{94.20\%} & \textbf{90.54\%} & \textbf{90.54\%} & \textbf{0.0\%} \\
REMIND & Custom Video & no background  & N/A & 80.78\% & 55.56\% & 55.56\% & \textbf{0.0\%} \\
REMIND & Custom Video & no neighbors   & N/A & 93.20\% & 69.73\% & 69.73\% & \textbf{0.0\%} \\

\hline

REMIND & ScanNet++ & base          & N/A & 67.32\% & \textbf{62.86\%} & \textbf{62.08\%} & \textbf{0.0\%}  \\
REMIND & ScanNet++ & greedy        & N/A & 65.77\% & 61.88\% & 60.73\% & 18.3\% \\
REMIND & ScanNet++ & no background & N/A & 47.18\% & 37.55\% & 37.51\% & 22.6\% \\
REMIND & ScanNet++ & no neighbors  & N/A & \textbf{69.56\%} & 60.45\% & 60.32\% & 19.0\% \\

\hline
\end{tabular}%
}

\end{table}
\section{Conclusions and Future Work}

We presented REMIND, a re-identification tracker purpose-built for long-term multi-object identity persistence of generic indoor objects from monocular video, with neither camera pose nor depth. By coupling frozen DINOv3 features with a dual-bank multi-prototype memory, part- and background-level descriptors, neighbour-context reasoning, and joint Hungarian assignment with ambiguity-aware guards, REMIND outperforms the strongest VOS baseline by nearly 20 IDF1 points on our custom indoor sequence, as well as a strong tracking-by-detection baseline (MASA) by more than 36; and retains the leading IDF1 on ScanNet++ in both forced-comparison regimes and under ground-truth masks on all scenes, ceding a marginal IDF1 lead to MASA only under end-to-end YOLO detection on all scenes, where it still holds higher association accuracy and recovery; it achieves this while completing 100\% of scenes against a 66.9\% out-of-memory failure rate of the VOS baseline under YOLO detections, suggesting that appearance memory and contextual reasoning, rather than explicit geometric reconstruction, are sufficient primitives for this setting.

Despite the success of the approach, several avenues remain open. Identity purity in dense same-class scenarios still dominates the residual error, so future improvements should target stronger same-class disambiguation and more discriminative descriptors rather than the recovery mechanism itself; a tighter detector (memory coupling, in which segmentation and re-identification share representations or feedback signals) should likewise narrow the gap to the ground-truth-mask regime. We further plan to extend the open-sourced custom video into a broader community benchmark spanning more environments, capture devices, and revisit patterns under a unified annotation protocol.

\section*{Author Contributions}
Conceptualization, P.D.-P. and A.R.-R.; methodology, P.D.-P. and A.R.-R.; software, P.D.-P.; validation, P.D.-P. and A.R.-R.; formal analysis, P.D.-P. and A.R.-R.; investigation, P.D.-P.; resources, P.C.; data curation, P.D.-P.; writing---original draft, P.D.-P. and A.R.-R.; writing---review and editing, P.D.-P., A.R.-R., D.P.-S. and P.C.; supervision, P.D.-P., A.R.-R., D.P.-S. and P.C.; project administration, A.R.-R. and P.C.; funding acquisition, P.C. All authors have read and agreed to the published version of the manuscript.

\section*{Acknowledgements}
This work has been supported by the project SHEREC ``Safe Healthy and Environmental Ship Recycling'', Reference: 101136056, funded by the European Union under the Horizon Europe Program HORIZON-CL4-2023-HUMAN-01 CNECT.

\bibliographystyle{IEEEtran}
\bibliography{references}

\appendices

\section{Threshold Parameters}
\label{sec:thresholds}

REMIND is a heuristic-driven system whose behavior is governed by a large collection of scalar thresholds, each introduced at its first use in the main text. Table~\ref{tab:thresholds} consolidates the most influential evaluation parameters, namely those whose values most directly shape the metrics reported in this work; the remaining internal coefficients, together with the experimental values adopted in this study, are documented in the provided open-source repository.

\begin{table}[H]
\renewcommand{\arraystretch}{1.15}
\setlength{\tabcolsep}{4pt}
\caption{REMIND Threshold Parameters}
\label{tab:thresholds}
\centering
\footnotesize
\begin{tabularx}{\columnwidth}{@{}p{1.9cm}lX@{}}
\hline
\textbf{Symbol} & \textbf{Stage} & \textbf{Description} \\
\hline
$\rho$                              & Perception   & Patch retention fraction for trimmed-mean descriptor \\
$K$                                 & Perception   & Number of K-means clusters for part descriptors \\
$\tau_{\mathrm{merge}}$             & Perception   & Cosine threshold for part cluster merging \\
$r_{\mathrm{in}}$                   & Perception   & Inner ring radius (patches) for background descriptor \\
$r_{\mathrm{out}}$                  & Perception   & Outer ring radius (patches) for background descriptor \\
\hline
$\tau_{\mathrm{dup}}$               & Memory       & Cosine threshold for work-bank duplicate detection \\
$\alpha$                            & Memory       & EMA step size for prototype update \\
$\tau_{\mathrm{merge,internal}}$    & Memory       & Cosine threshold for work-prototype internal merging \\
$h_{\mathrm{confirm}}$              & Memory       & Hit count for \textsc{Tentative}$\to$\textsc{Confirmed} transition \\
$m_{\mathrm{max}}$                  & Memory       & Max.\ consecutive misses before \textsc{Confirmed}$\to$\textsc{Inactive} \\
$\mathrm{TTL}$                      & Memory       & Frames before \textsc{Inactive}/ambiguous tracks are removed \\
\hline
$w_{\mathrm{obj}}, w_{\mathrm{bg}},$ \newline $w_{\mathrm{parts}}$ & Association  & Initial nominal weights for object, background, and part channels (adjusted per-frame by quality factors $q_c$) \\
$\tau_{\mathrm{match}}$             & Association  & Minimum similarity score for a detection to be assigned to an existing track (most critical parameter) \\
$\tau_{\mathrm{vis,min}}$           & Association  & Minimum visual score below which contextual support may rescue an otherwise sub-threshold match \\
$\delta_{+}$                        & Association  & Max.\ contextual support bonus per candidate \\
$\delta_{-}$                        & Association  & Max.\ contextual contradiction penalty per candidate \\
$\delta_{\mathrm{confirm}}$         & Association  & Min.\ score margin (best vs.\ second-best) for anchor qualification \\
$\tau_{\mathrm{lock}}$              & Association  & Score threshold for high-confidence pre-assignment \\
\hline
\end{tabularx}
\end{table}

For completeness, Table~\ref{tab:masa_settings} summarises the evaluation-specific MASA settings used in the comparisons. These adaptations build on the official MASA-GDINO setup and align the baseline with the long-term indoor re-identification setting evaluated in this work.

\begin{table}[H]
\renewcommand{\arraystretch}{1.15}
\setlength{\tabcolsep}{4pt}
\caption{MASA evaluation settings}
\label{tab:masa_settings}
\centering
\footnotesize
\begin{tabularx}{\columnwidth}{@{}p{2.3cm}X@{}}
\hline
\textbf{Setting} & \textbf{Value / Description} \\
\hline
Base model & Official MASA-GDINO configuration \\
Class-aware association & Enabled; candidate matches are restricted to instances of the same semantic class \\
Spatial distance filtering & Disabled, to avoid penalising large viewpoint changes and long temporal gaps \\
Memory duration & Full-scene memory, retained across disappearances and revisits \\
Embedding momentum & $0.3$, selected as the best empirical value among the tested settings \\
\hline
\end{tabularx}
\end{table}




\end{document}